\documentclass[10pt,twocolumn,letterpaper]{article}

\usepackage{cvpr}
\usepackage{times}
\usepackage{epsfig}
\usepackage{graphicx}
\usepackage{amsmath}
\usepackage{amssymb}

\newcommand{\gmh}[1]{\textcolor[rgb]{0, 0, 0}{#1}}

\usepackage[breaklinks=true,bookmarks=false]{hyperref}

\cvprfinalcopy 

\def\cvprPaperID{****} 

\begin{document}

\title{Can Attention Enable MLPs To Catch Up With CNNs?}

\newcommand*{\affaddr}[1]{#1} 
\newcommand*{\affmark}[1][*]{\textsuperscript{#1}}
\newcommand*{\email}[1]{\texttt{#1}}

\author{
Meng-Hao Guo\affmark[1], Zheng-Ning Liu\affmark[1], Tai-Jiang Mu\affmark[1], Dun Liang\affmark[1], \\ Ralph R. Martin\affmark[2] and Shi-Min Hu\affmark[1]\\
\affaddr{\affmark[1]BNRist, Department of Computer Science and Technology, Tsinghua University, Beijing}\\
\affaddr{\affmark[2]Cardiff University}\\
{\{gmh20, liu-zn17, liangd16\}@mails.tsinghua.edu.cn,} \\
  {\{taijiang, shimin\}@tsinghua.edu.cn, ralph@cs.cf.ac.uk}
}

\maketitle
\begin{abstract}
    
In the first week of May, 2021, researchers from four different  institutions: Google, Tsinghua University, Oxford University and Facebook, shared their latest work~\cite{Tolstikhin2021MLP-Mixer,guo2021attention,melaskyriazi2021need,touvron2021resmlp} on \url{arXiv.org} almost at the same time, each proposing new learning architectures,  consisting mainly of linear layers, claiming them to be comparable, or even superior to convolutional-based models.
This sparked immediate  discussion and debate in both academic and industrial communities as to whether MLPs are sufficient, many  thinking that learning architectures are returning to MLPs. 
Is this true? 

In this perspective, we give a brief history of learning architectures, including multilayer perceptrons (MLPs), convolutional neural networks (CNNs) and  transformers. We then examine what the four newly proposed architectures  have in common. Finally, we  give our views on challenges and directions for new learning architectures, hoping to inspire future research. 

\end{abstract}

\section{Learning architectures for visual tasks}\label{sec:introduction}

Multilayer perceptrons (MLPs)~\cite{hinton_mlp} consist of an input layer and an output layer, possibly with multiple hidden layers in between.
Layers are typically fully connected with linear transformations and activation functions.
MLPs were the basis for neural networks before deep convolutional neural networks (DCNNs) took over, and  greatly improved the power of computers to handle problems of classification and regression.
However, MLPs are computationally costly and prone to overfitting, due to their large numbers of parameters.
MLPs are also poor at capturing local structures in the input, since the linear transformations between layers  always take the output from the previous layer as a whole.
However, we note that the capabilities of MLPs were not fully explored when they were proposed, both because of limited computer performance,  and unavailability of massive data for training.

To learn local structures in the input while maintaining computational efficiency, convolutional neural networks (CNNs) were proposed.
In 1998, LeCun {\etal} presented LeNet~\cite{Lecun98LeNet}, which greatly improved the accuracy of handwritten digit recognition using a five-layer convolutional neural network. 
Later, AlexNet~\cite{Krizhevsky12AlexNet} lead to wide acceptance of CNNs  in the research community: it was much larger than previous CNNs like LeNet, and beat all other competitors by a significant margin in the ImageNet Large Scale Visual Recognition Challenge in 2012\footnote{https://image-net.org/challenges/LSVRC/2012/}.
Since then, many more models with ever deeper architectures have been developed, with many providing more accurate results than humans in many realms, resulting in profound paradigm changes in both scientific research, and engineering and commercial applications.

Putting aside the advances in computing power and amounts of training data, the key success of CNNs lies in the inductive bias they introduce: they assume that information has spatial locality and can thus reduce the number of network parameters by making use of a sliding convolution with shared weights.
However, the side-effect of this approach is that the receptive fields of CNNs are limited, making CNNs less able to learn long-range dependencies.
To enlarge the receptive field, a larger convolutional kernel is required, or other special strategies must be employed, such as dilated convolutions~\cite{yu2016multiscale}.
Note that composing a large kernel from several small kernels is not  a suitable approach for enlarging the receptive field of CNNs~\cite{Peng17LargeKernel}. 

Recently, the Transformer neural network architecture was proposed~\cite{Vaswani17Transformer} for sequential data, with great success in natural language processing~\cite{Radford18GPT,Devlin19BERT}, and more recently, vision~\cite{dosovitskiy2021ViT,carion2020endtoend,Guo21pct,yuan2021tokenstotoken, touvron2021training}.
The attention mechanism is at the core of Transformer, which is readily capable of  learning long-range dependencies between any two positions in the input data in the form of an attention map.
However, this additional freedom and reduced inductive bias mean that  effectively training Transformer-based architectures requires huge amounts of data. 
For best results, such models should be first pre-trained on a very large dataset, such as GPT-3~\cite{Brown20GPT-3} and ViT~\cite{dosovitskiy2021ViT}.



\section{Linear Layer Based Architectures}

\subsection{Four Recent Architectures}

To avoid the drawbacks of the aforementioned learning architectures, and, with the aim of achieving better results at lower computational cost, very recently, four architectures were proposed almost simultaneously~\cite{Tolstikhin2021MLP-Mixer,guo2021attention,melaskyriazi2021need,touvron2021resmlp}. Their common aim is to take full advantage of linear layers. 
\begin{figure*}
    \centering
    \includegraphics[width=0.71\textwidth]{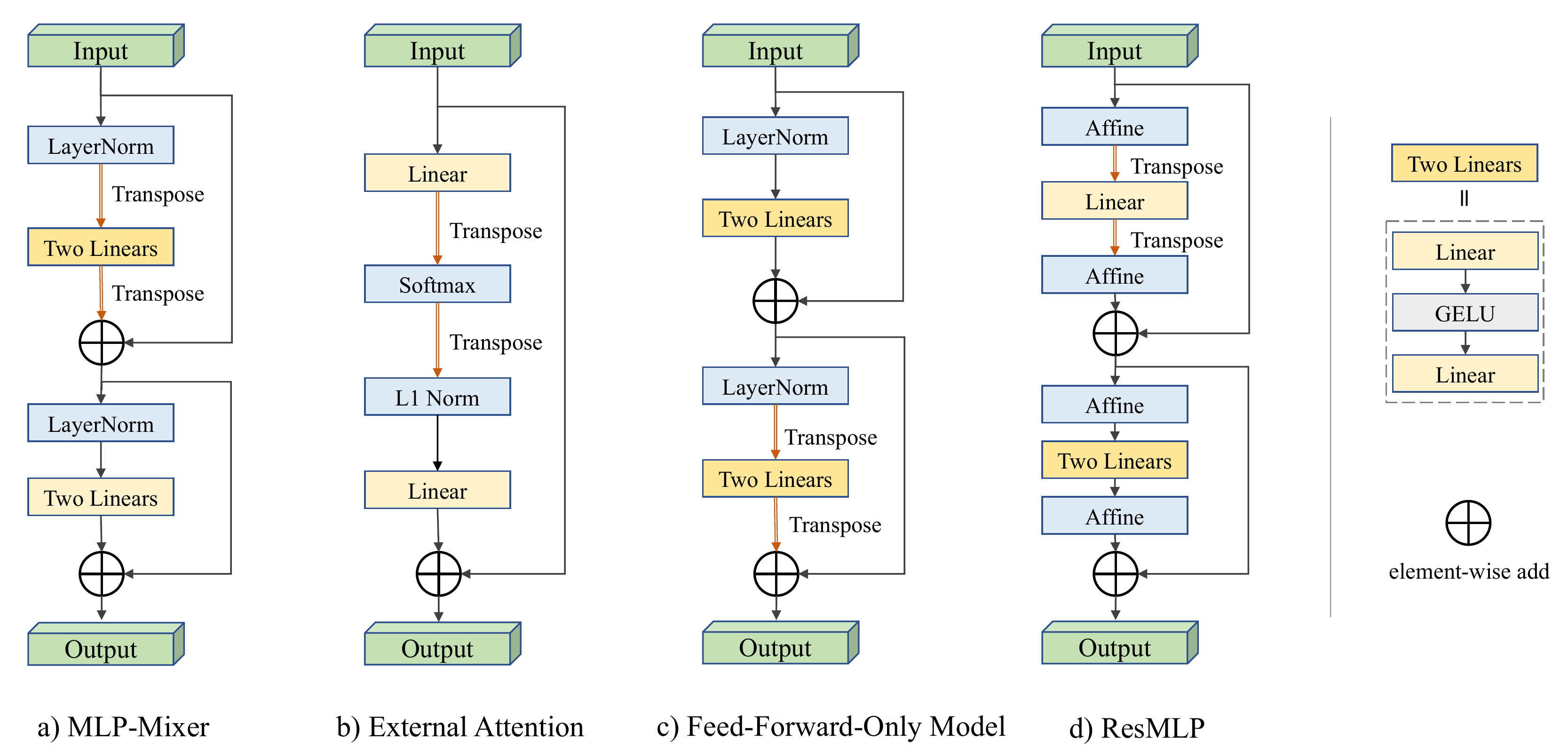}
    \caption{Four recent architectures in which linear layers predominate. }
    \label{fig:my_label}
\end{figure*}
We  briefly summarize these architectures below; also see Fig.~\ref{fig:my_label}. All four employ transposition to model interactions at all scales. Residual connections and normalization are utilized in a similar way to ensure stable training.   

\subsection{MLP-Mixer}
MLP-Mixer~\cite{Tolstikhin2021MLP-Mixer} takes $S$ non-overlapping image patches of resolution $P \times P$ as input. Each patch is first projected to a $C$-dimensional embedding via a shared-weight linear layer: this  representation of the input image is thus a matrix, $\mathbf{X} \in \mathbb{R}^{S \times C}$. 

Next, $\mathbf{X}$ is fed into a sequence of identical \emph{mixer layers}, each of which is composed of a \emph{token-mixing} MLP block and a \emph{channel-mixing} MLP block, mixing  information from all patches, and from all channels, respectively. We may express the computation as:
\begin{align}
    \mathbf{U} & = \mathbf{X} + f_2 (\sigma(f_1 (\text{Norm}(\mathbf{X})^T)))^T, \\
    \mathbf{Y} & = \mathbf{U} + f_4 (\sigma(f_3 (\text{Norm}(\mathbf{U}))), 
\end{align}
where $f_1, \cdots, f_4$ are linear layers, and $\sigma$ denotes GELU (nonlinear) activation~\cite{hendrycks2016gaussian}. Layer normalization~\cite{ba2016layer} is employed. $\mathbf{U} \in \mathbb{R}^{S \times C}$ is the intermediate matrix after  per-channel feature aggregation: a shared-weight mapping $\mathbb{R}^S \mapsto \mathbb{R}^S$ of the column vectors in $\mathbf{X}$. Similarly, two linear transformations are performed per patch, giving the output $\mathbf{Y}$.

\subsection{External Attention}
 External attention~\cite{guo2021attention} \gmh{reveals} the relation between self-attention and linear layers. It first simplifies self-attention as in Eq.~\ref{eq.simplified_attention}, where $\mathbf{M} \in \mathbb{R}^{N \times d}$ is the input feature map. 
\begin{align}
    \mathbf{A} & = \mathrm{softmax}(\mathbf{F} \mathbf{F}^T), \\
    \mathbf{F_{out}} & = \mathbf{A}\mathbf{F}.
\label{eq.simplified_attention}
\end{align}
Then an external memory unit $\mathbf{M} \in \mathbb{R}^{S \times d}$ is introduced to replace $\mathbf{M}$-to-$\mathbf{M}$ attention by $\mathbf{M}$-to-$\mathbf{M}$ attention as below:
\begin{align}
    \mathbf{A} & = (\alpha)_{i,j} = \mathrm{Norm}(\mathbf{F} \mathbf{M^T}), \\
    \mathbf{F_{out}} & = \mathbf{A} \mathbf{M}.
\label{eq.f-to-m}
\end{align}
Finally, like self-attention, it uses two different memory units $M_k$ and $M_v$ as the key and the value to increase the capability of the network. The overall computation of external attention is as below:
\begin{align}
    \mathbf{A} & = \mathrm{Norm}(\mathbf{F} \mathbf{M_k^T}), \\
    \mathbf{F_{out}} & = \mathbf{A} \mathbf{M_v}.
\label{eq.final_attention_1}
\end{align}
Because $\mathbf{F} \mathbf{M_k^T}$ is matrix multiplication, it is linear in $\mathbf{F}$, so Eq.~\ref{eq.final_attention_1} can be written as
\begin{align}
    \mathbf{F_{out}} & = f_2 (\text{Norm}(f_1(\mathbf{F}))).
\label{eq.final_attention_2}
\end{align}

The final output is then obtained by adding an identity mapping as below:
\begin{align}
    \mathbf{F_{out}} & = \mathbf{F} + f_2 (\text{Norm}(f_1(\mathbf{F}))). 
\label{eq.final_attention}
\end{align}

\gmh{Based on the external attention, Guo et al.~\cite{guo2021attention} also design a multi-head external attention and achieve an all MLP architecture named EAMLP.}

\subsection{Feed-forward-only Model}

The feed-forward-only model~\cite{melaskyriazi2021need} replaces the attention layers in Transformer~\cite{Vaswani17Transformer} by simple feed-forward layers on the token dimension. It firstly uses linear layers on the channel dimension and then adopts linear layers on the token dimension in a linear block. Given an input $\mathbf{X} \in \mathbb{R}^{N \times C}$, the computation in detail can be expressed as:
\begin{align}
    \mathbf{U} & = \mathbf{X} + f_2 (\sigma(f_1 (\text{LayerNorm}(\mathbf{X})))), \\
    \mathbf{Y} & = \mathbf{U} + f_4 (\sigma(f_3 \text{LayerNorm}(\mathbf{U}^T)))^T.
\label{eq.linearblock}
\end{align}

\subsection{ResMLP}
ResMLP~\cite{touvron2021resmlp} also separately aggregates information in  per-patch-style and per-channel-style, and can be formulated as  follows:
\begin{align}
    \mathbf{U} & = \mathbf{X} + \text{Norm}(f_1 (\text{Norm}(\mathbf{X})^T)^T) \\
    \mathbf{Y} & = \mathbf{U} + \text{Norm}(f_3 (\sigma(f_2( \text{Norm}(\mathbf{U})))) 
\end{align}

A major difference of ResMLP is that it uses an affine transformation in the role of a normalization layer. This affine transformation is parameterized by two learnable vectors to scale and shift the input component-wise:
\begin{equation}
    \text{Aff}_{\mathbf{\alpha}, \mathbf{\beta}} (\mathbf{X}) = \text{Diag}(\mathbf{\alpha}) \mathbf{x} + \mathbf{\beta}
\end{equation}
Note that no statistics of the input are used in the above, and thus it can be integrated in the linear layers during inference for further speed.

\section{Common Themes}

We now examine the above approaches, to see what they have in common. 

\subsection{Long distance interactions}
As in self-attention,  interactions between different patches are taken into account by these four methods. MLP-Mixer, ResMLP and the Feed-forward-only model use linear layers acting on the token dimension to allow different patches to communicate with each other. External attention adopts softmax and $L1$ normalization to perform a similar role. Unlike CNNs, these models can consider long distance interactions between patches and automatically select suitable and irregular receptive fields. 

\subsection{Local semantic information}
Unlike independent words in natural language, single pixels have very little semantic information and their interactions with other pixels are not directly informative. It is thus important to extract meaningful information before using MLPs. MLP-Mixer, ResMLP and the Feed-forward-only model divide the image into $16 \times 16$ local patches to obtain semantic information. External attention adopts a \gmh{T2T module~\cite{yuan2021tokenstotoken} or} CNN backbone to provide rich semantics before passing information to linear layers.

\subsection{Residual connections}
Residual connections~\cite{resnet} solve the problem of vanishing gradients and stabilize the training process, so they are commonly used in deep convolutional neural networks. They also benefit architectures based around linear layers and are adopted by all the above models. 

\subsection{Reduced inductive bias}
Localised processing in CNNs results in inductive bias, which can decrease accuracy when the training data is sufficient. The recently introduced architectures use linear layers on single tokens independently, or process all tokens equally, resulting in lower inductive bias than CNNs.

\section{Challenges and future directions}

These promising recently introduced architectures have simple network structure and fast inferencing throughput. However, on ImageNet, their results are currently 5--10\% less accurate than those provided by the best CNNs or Transformer networks. They also do not significantly outperform light-weight networks in the trade-off between accuracy and speed.  Thus additional research is needed if the potential of such architectures is to be realised.  

We suggest possible directions for future work below, and make other observations about these architectures:
\begin{itemize}

\item All linear layers process image patches in a direct or indirect manner, to extract local features thereby reducing computational cost. Dividing images into non-overlapping patches again introduces  inductive bias. On one hand, CNNs capture local structure extremely well, but lack the ability to handle long range interactions. On the other hand, these four architectures provide a good way to process long range interactions. It seems natural to try to combine the advantages of both architectures.

\item One main goal of these four methods is to avoid the use of the self-attention mechanism. The successful configurations used for this purpose in Transformer could be employed in these linear architectures. For example, Transformer can use multi-head attention, and a similar multi-head mechanism could be employed by these methods to improve model capability.

\item Residual connections play a key role in all these methods, indicating that the network structure is crucial. Because these new architectures are simpler than CNNs, better backbones are needed.

\item Due to the simplicity of these new architectures, they can easily tackle irregular data structures, including point clouds, graphs, etc., used  in various applications. Furthermore, this flexibility promises the ability to make cross-modal models, with a unified network backbone for all modes of data. 

\item An additional benefit is that all computations are matrix multiplications, which can be highly optimized in deep learning frameworks and readily performed on hardware. This simplicity can promote deployment in industry and commerce, and also reduce  energy consumption. 

\end{itemize}

\section{Conclusions} 
Overall, the new architectures separately apply linear layers in the element (token) dimension and channel dimension to learn long range interactions between any two positions in the feature matrix, while traditional MLPs mix these two dimensions together as a long vector, with  too much freedom for effective learning. 
We conclude that the new  architectures do not simply reuse traditional MLPs, but are a significant advance over them.

{\small
\bibliographystyle{ieee}
\bibliography{egbib}
}

\end{document}